\documentclass{article}

\usepackage{arxiv}

\usepackage[utf8]{inputenc} % allow utf-8 input
\usepackage[T1]{fontenc}    % use 8-bit T1 fonts
\usepackage{hyperref}       % hyperlinks
\usepackage{url}            % simple URL typesetting
\usepackage{booktabs}       % professional-quality tables
\usepackage{amsfonts}       % blackboard math symbols
\usepackage{nicefrac}       % compact symbols for 1/2, etc.
\usepackage{microtype}      % microtypography
\usepackage{lipsum}		% Can be removed after putting your text content
\usepackage{graphicx}
\usepackage{natbib}
\usepackage{doi}
\usepackage{amsmath}

\title{Advances in Machine Learning Research Using Knowledge Graphs}

\date{} 					% Or removing it

\author{{\includegraphics[scale=0.06]{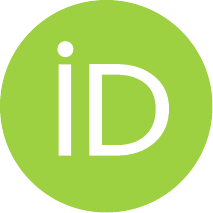}\hspace{1mm}Jing~Si}\thanks{These authors contributed equally to this work. Funding Project: Ministry of Education Humanities and Social Sciences Research Youth Fund Project; Project Number: 15YJCZH232; Shanghai Science and Technology Committee Soft Science Project; Project Number: 17692109200; Shanghai College Students Innovation Training Program; Project Number: cs1703008.} \\
	Department of Management\\
	Shanghai University of Engineering Science\\
	Shanghai, China 201620 \\
	\texttt{1135905683@qq.com} \\
        \And
	\href{https://orcid.org/0009-0006-3794-2802}{\includegraphics[scale=0.06]{orcid.pdf}\hspace{1mm}Jianfei~Xu\footnotemark[1]} \\
	Department of Management\\
	Shanghai University of Engineering Science\\
	Shanghai, China 201620 \\
	\texttt{mr.xujianfei@gmail.com} \\
}

\renewcommand{\headeright}{}
\renewcommand{\undertitle}{}
\renewcommand{\shorttitle}{}

\hypersetup{
pdftitle={Advances in Machine Learning Research Using Knowledge Graphs},
pdfsubject={cs.AI},
pdfauthor={Jianfei Xu, Jing Si},
pdfkeywords={Machine Learning, Knowledge Graph, Visual Analytics},
}

\begin{document}
\maketitle

\begin{abstract}

The study uses CSSCI-indexed literature from the China National Knowledge Infrastructure (CNKI) database as the data source. It utilizes the CiteSpace visualization software to draw knowledge graphs on aspects such as institutional collaboration and keyword co-occurrence. This analysis provides insights into the current state of research and emerging trends in the field of machine learning in China. Additionally, it identifies the challenges faced in the field of machine learning research and offers suggestions that could serve as valuable references for future research.

\end{abstract}

% keywords can be removed

\section{Introduction}

Machine learning is an interdisciplinary field that studies how computers can learn and simulate human learning behaviour. By acquiring new knowledge, machine learning aims to reorganize existing knowledge structures to continuously improve its own performance. Machine learning was proposed in the mid-1950s, and over the next 30 years, related research in the field of machine learning continued to develop. Machine learning has interdisciplinary attributes and has been widely applied in the field of artificial intelligence. \cite{zhang2016} argue that the way to transform big data into more valuable knowledge is by applying machine learning techniques. Through empirical research, they concluded that, to significantly improve the performance of machine learning models, the scale of data needs to be expanded. \cite{guo2010} state that the core of machine learning is learning, and the focus of research is how to enable machines to improve their performance in a way similar to humans, influenced by the external environment.

Many scholars have conducted related research on machine learning algorithms: \cite{he2014} explored six machine learning algorithms in the context of big data, including big data divide-and-conquer strategies, sampling, feature selection, classification, clustering, association analysis, and parallel algorithms. \cite{liu2017} discussed the issues and limitations of traditional machine learning and proposed parallel machine learning algorithms that are more suitable for handling large-scale, highly scalable data storage in big data environments. \cite{xiao2017} provided corresponding suggestions and methods for machine learning algorithms under big data, based on research into the current status, characteristics, and classification of big data in China. \cite{cao2017} discussed the fields involved in current machine learning and its roles, affirming that machine learning plays a significant role in advancing artificial intelligence, and predicted the future development of machine learning based on existing achievements.

Currently, research in the field of machine learning in China is becoming increasingly popular, and its application scope is expanding. By studying the current state of machine learning in China, it will have significant implications for the future development trends and directions of machine learning in the country. This paper uses CSSCI papers in the field of machine learning in China as the data source and presents and evaluates the research status and hot topics of machine learning through the form of knowledge graphs.

\section{Data and Methods}

\subsection{Data Source}

In order to scientifically study the current state and trends of development in the field of machine learning, the author uses the China National Knowledge Infrastructure (CNKI) database as the data source. The search term "machine learning" is used as the title, with the literature classified under "Information Technology". The time range for the publication of the papers is set from 2007 to 2017, and the search scope is limited to CSSCI journals.

\subsection{Research Methods and Tools}

With the advancement of technology, visualization analysis software has become a cutting-edge method for analyzing bibliometric information, providing intuitive visual processing and analysis of literature. CiteSpace is the primary analysis software used in this paper. Developed by scholars on the Java platform, CiteSpace is a highly visual tool. It has the ability to display, identify, and analyze the current status and development prospects of relevant research fields, making it a powerful visualization tool for analyzing scientific literature data \cite{lin2017}.

\section{Bibliometric Analysis}

\begin{figure}[htbp]
    \centering
    \includegraphics[width=0.5\linewidth]{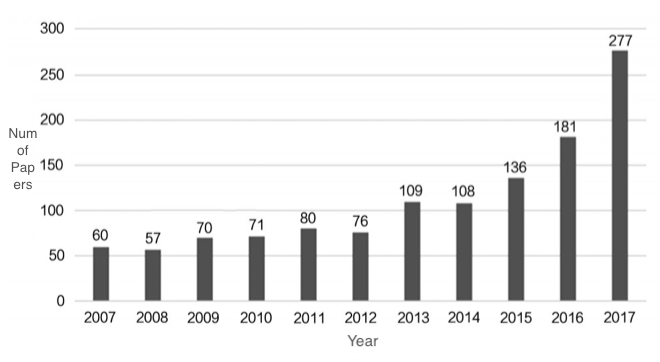}
    \caption{Trends in the Volume of Machine Learning Literature}
    \label{fig:trends_in_the_volume_of_machine_learning_literature}
\end{figure}

This paper analyzes the research status of the field of machine learning over the past decade by compiling the number of CSSCI papers published each year, as shown in Figure \ref{fig:trends_in_the_volume_of_machine_learning_literature}. It is evident that from 2007 to 2012, the number of papers related to machine learning research in China showed a gradual increase, indicating that during this period, there were no groundbreaking advances in the field, and the research was progressing slowly. After 2014, however, the number of published papers surged sharply, marking this period as the outbreak of the research boom in domestic machine learning studies. Subsequently, the field's output grew rapidly. Particularly after the 2016 victory of AlphaGo over Lee Sedol in the game of Go, artificial intelligence once again gained significant attention from scholars. During the period of 2016 to 2017, the focus and importance of machine learning research increased rapidly, becoming a major research hotspot.

\subsection{Institutional Distribution}

\begin{figure}
    \centering
    \includegraphics[width=0.5\linewidth]{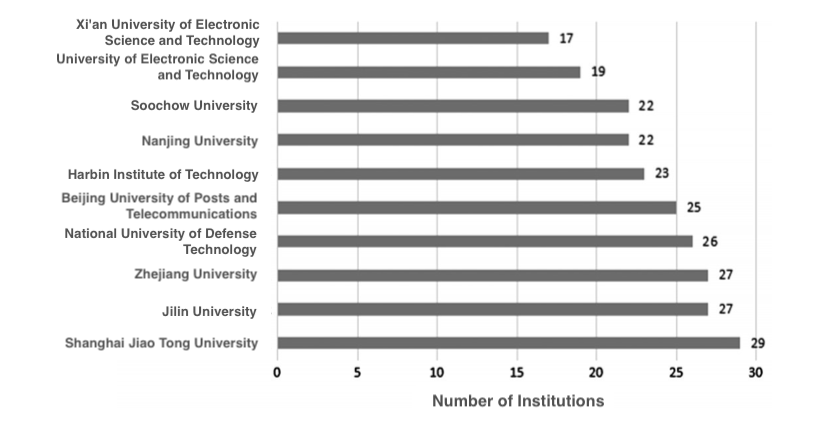}
    \caption{High-Output Institutions for Machine Learning Research}
    \label{fig:high_output_institutions_for_machine_learning_research}
\end{figure}

The research status of machine learning in China is analyzed from the perspective of institutions and academic teams. Using CiteSpace, high-output research institutions were identified, as shown in Figure \ref{fig:high_output_institutions_for_machine_learning_research}. From the perspective of published literature, Shanghai Jiao Tong University, Jilin University, and Zhejiang University are the leading institutions in machine learning research. This indicates that these three research institutions have strong influence and research capabilities in the field of machine learning. Following closely behind are seven other research institutions, including the National University of Defense Technology, Beijing University of Posts and Telecommunications, and Harbin Institute of Technology.

To analyze the collaboration between research institutions, relevant parameters were set in CiteSpace. The thresholds for C/CC/CCV were set as (2, 2, 20), (4, 3, 20), and (4, 3, 20), respectively. Additional settings included Top N=30, Top N\%=15, and Article Labeling Threshold=2. This resulted in the collaborative network map of machine learning research institutions, as shown in Figure \ref{}. The font size of the labels represents the strength of centrality—the larger the font size, the stronger the centrality. Node rings represent annual layers, while edges indicate collaboration relationships between institutions. The results show that the number of network nodes is N=23, the number of edges is E=3, and the network density is Density=0.0119. This indicates that collaboration between domestic institutions in the field of machine learning research is relatively limited, and the connections between institutions are not sufficiently close.

An analysis was conducted on the research activities of various institutions across different time windows, and a temporal map of institutions in the field of machine learning research was generated, as shown in Figure \ref{}. Different color intervals represent different time periods, and the node colors correspond to the time period colors above. In 2007, the School of Computer Science and Technology at Soochow University published machine learning-related research results in core journals. The analysis results indicate that this institution has had a significant impact, and its publication volume has shown a continuous growth trend in subsequent years.

\subsection{Author Distribution}

\begin{figure}[htbp]
    \centering
    \includegraphics[width=0.5\linewidth]{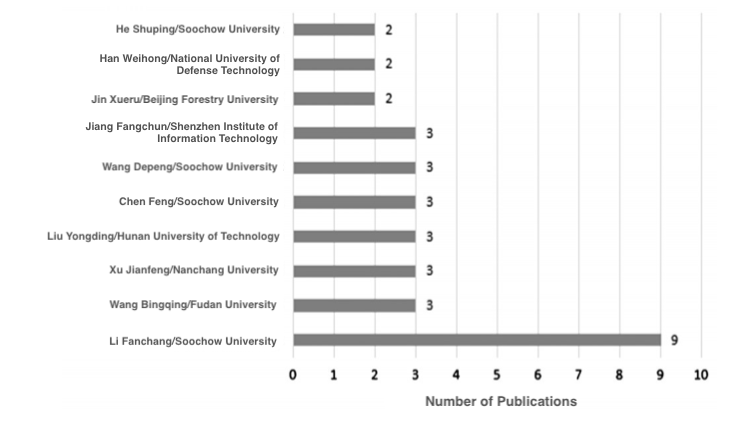}
    \caption{Prolific Authors and Publication Volume}
    \label{fig:prolific_authors_and_publication_volume}
\end{figure}

An analysis of the distribution of authors in the field of machine learning was conducted. Among all the retrieved documents, a total of 46 authors were identified. Based on the bibliometric law of prolific authors, the top ten authors by publication volume were selected for statistical analysis, as shown in Figure \ref{fig:prolific_authors_and_publication_volume}. Notably, 2/5 of the prolific authors come from Soochow University, once again proving the scientific research level and strength of this research team in the field. It is also worth noting that scholars from Fudan University, Nanchang University, and Hunan University of Technology are closely following, with significant publication volumes as well.

\begin{figure}
    \centering
    \includegraphics[width=0.5\linewidth]{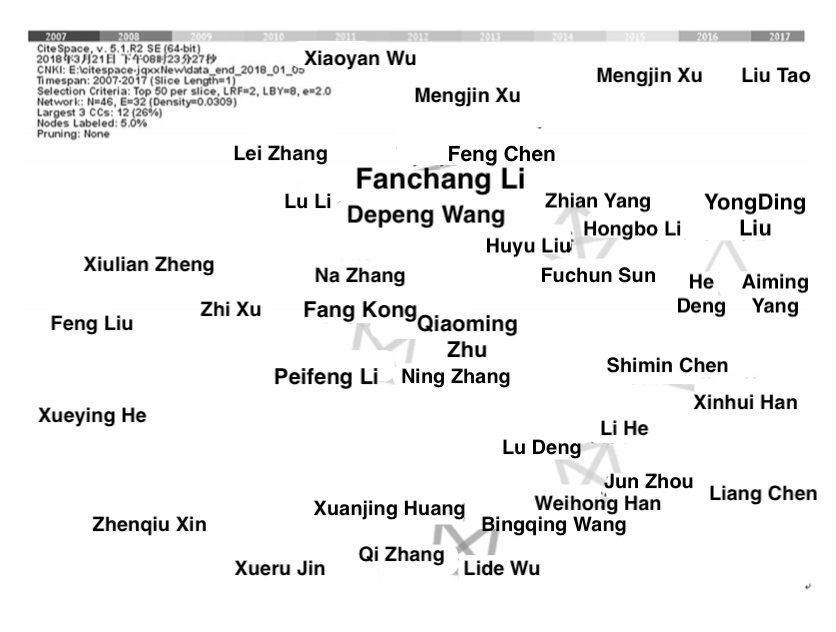}
    \caption{Author Collaboration Map}
    \label{fig:author_collaboration_map}
\end{figure}

An analysis of author collaboration was conducted, with the CiteSpace parameter settings consistent with those previously described. The generated collaboration network map for machine learning research institutions is shown in Figure \ref{fig:author_collaboration_map}. The font size of the labels represents the strength of centrality—the larger the font size, the stronger the centrality. Node rings indicate annual layers, while edges reflect collaboration relationships between institutions.

The results reveal that the network consists of \( N=46 \) nodes, \( E=32 \) edges, and a network density of \( Density=0.0372 \). Considering that there are as many as 26 institutions involved in machine learning research, this network density is relatively low. This indicates that collaboration among domestic authors is rare, with a general lack of collaborative awareness. Moreover, the limited collaboration that does occur is often confined within the same research institution, rather than forming robust cross-institutional research groups.

\subsection{Keyword Co-Occurrence Ranking}

Keywords typically represent the key focus areas of research within a specific field. By conducting a co-occurrence analysis of keywords, it is possible to accurately grasp the current state and research hotspots of a field. The concept of keyword co-occurrence originates from bibliometrics, specifically citation analysis and bibliographic coupling theory, which is used to analyze the strength of connections between documents \cite{wen2016}.

\begin{figure}
    \centering
    \includegraphics[width=0.5\linewidth]{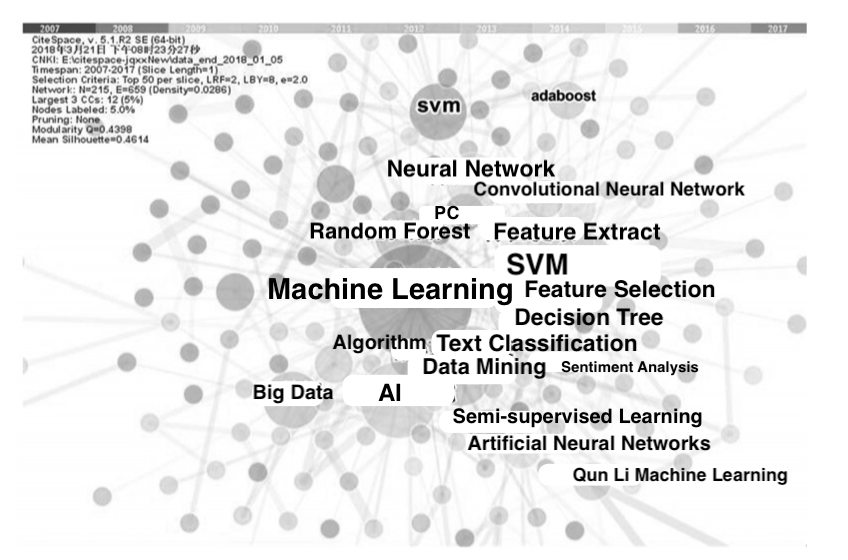}
    \caption{Keyword Co-Occurrence Knowledge Map}
    \label{fig:keyword_co-occurrence_knowledge_map}
\end{figure}

In CiteSpace, the "Keyword" node type was selected, with parameter settings consistent with those previously described. The generated keyword co-occurrence knowledge map is shown in Figure \ref{fig:keyword_co-occurrence_knowledge_map}. The map reveals a total of 235 nodes and 754 connections.

The co-occurrence relationships between keywords are reflected through the edges connecting nodes in the figure \ref{}. The size of a node indicates the frequency of the corresponding keyword—larger nodes represent higher frequencies. The top ten keywords by frequency are listed in Table 1, which also records the centrality of each keyword. Overall, research hotspots in this field are relatively dispersed, with a low network density for keyword co-occurrence, indicating no clear concentration of research focus. Since all the analyzed literature is related to "machine learning," the term "machine learning" itself was excluded from Table \ref{}. From the statistical results, it can be observed that machine learning research is closely related to fields or algorithms such as support vector machines, neural networks, and data mining. Scholars place significant emphasis on support vector machines (SVM), which are an important supervised learning model in machine learning. SVMs are commonly used for classification and regression analysis to identify and analyze data. They can be flexibly applied to practical scenarios involving high-dimensional and non-linear problems and are grounded in a rigorous theoretical foundation \cite{chen2016}. Neural networks, on the other hand, are computational models inspired by the brain's mechanisms to analyze data \cite{zhang2017}. With the explosion of information, machine learning today primarily operates in big data environments to conduct corresponding analyses and research. The foundation of big data is data mining \cite{jiang2016}, which is inherently linked to machine learning. Data mining involves using relevant algorithms to extract hidden information from massive, complex datasets \cite{li2014}. Table \ref{tab:explanation_and_description_of_variables} shows that these keywords exhibit high centrality in the co-occurrence network graph, leading to the preliminary conclusion that support vector machines, feature selection, neural networks, artificial intelligence, and data mining are research hotspots in the field of machine learning.

\begin{table}[htbp]
\centering
\begin{tabular}{lccc}
\toprule
\textbf{Rank} & \textbf{Keyword}         & \textbf{Frequency} & \textbf{Centrality} \\ 
\midrule
1             & Support Vector Machine   & 145                & 0.22                \\ 
2             & Feature Selection        & 49                 & 0.11                \\ 
3             & Neural Network           & 42                 & 0.07                \\ 
4             & Artificial Intelligence  & 40                 & 0.03                \\ 
5             & Data Mining              & 39                 & 0.09                \\ 
6             & Decision Tree            & 33                 & 0.04                \\ 
7             & Genetic Algorithm        & 33                 & 0.05                \\ 
8             & Feature Extraction       & 33                 & 0.13                \\ 
9             & Supervised Learning      & 31                 & 0.13                \\ 
10            & Pattern Recognition      & 30                 & 0.10                \\ 
\bottomrule
\end{tabular}
\caption{Explanation and Description of Variables}
\label{tab:explanation_and_description_of_variables}
\end{table}

\section{Conclusion and Recommendations}

This study utilizes CiteSpace visualization software to analyze machine learning literature cited in China's CSSCI from 2007 to 2017 by generating knowledge maps. The conclusions and recommendations based on the findings are as follows:

\begin{itemize}
    \item Temporal Distribution Map. The temporal distribution map shows that research outcomes in machine learning have developed rapidly over the past decade, particularly after 2014, when the growth rate increased significantly. Currently, the field is experiencing a phase of rapid growth, with research results primarily published in computer and information-related journals. At this stage, the importance of the field is gradually increasing, with trends indicating further expansion in both the breadth and depth of research.
    \item Research Institution and Author Distribution. The distribution maps of research institutions and authors reveal that numerous institutions are involved in the field of machine learning, presenting a diversified landscape. Strong research teams have emerged, such as those from Soochow University, Shanghai Jiao Tong University, Jilin University, and Zhejiang University. However, collaboration between institutions remains insufficient. While prolific authors have contributed significantly to the field's development, their collaboration is mostly limited within their own institutions, with limited cross-institutional exchanges. Additionally, some authors have followed trends blindly, leading to overlapping research outcomes and a lack of research depth. Research institutions should ensure that increased funding for research is accompanied by efforts to maintain researchers' enthusiasm and focus. Organizing research exchange activities could help establish a stable and positive collaborative environment to collectively advance the field.
    \item Keyword Co-Occurrence Map. The keyword co-occurrence map reveals that related algorithms such as support vector machines, neural networks, and data mining are widely emphasized and discussed, attracting significant attention. However, from a broader perspective, the keyword structure of machine learning research remains relatively dispersed, indicating that research hotspots have yet to converge.
    \item Co-Citation Information. The co-citation analysis shows that the most frequently cited journals and papers in the field of machine learning are concentrated in information and computer-related journals, highlighting the technical nature of the field. The high proportion of research published in information-related journals demonstrates the importance attached to this field and the strong support for the publication of its research outcomes.
    \item Research Frontiers Temporal Map. The temporal map of research frontiers indicates that the field of machine learning covers a wide range of topics and exhibits significant interdisciplinary characteristics. The rise of artificial intelligence has accelerated the development of algorithms such as random forests and convolutional neural networks in machine learning. Additionally, hot topics such as "sentiment classification" and "big data" have started to emerge in the field of machine learning, driving the research in this area to deeper levels.
\end{itemize}

These findings provide a comprehensive overview of the development, collaboration, and research trends in the field of machine learning, offering valuable insights for future research directions and strategies.
\newpage
\bibliographystyle{unsrtnat}
\bibliography{references} 

\begin{thebibliography}{12}
\providecommand{\natexlab}[1]{#1}
\providecommand{\url}[1]{\texttt{#1}}
\expandafter\ifx\csname urlstyle\endcsname\relax
  \providecommand{\doi}[1]{doi: #1}\else
  \providecommand{\doi}{doi: \begingroup \urlstyle{rm}\Url}\fi

\bibitem[Zhang and Wang(2016)]{zhang2016}
Run Zhang and Yongbin Wang.
\newblock Research on machine learning, its algorithms, and development.
\newblock \emph{Journal of Communication University of China (Natural Science Edition)}, \penalty0 (2):\penalty0 10--18, 24, 2016.

\bibitem[Guo and Feng(2010)]{guo2010}
Yaning Guo and Shasha Feng.
\newblock Theoretical research on machine learning.
\newblock \emph{China Science and Technology Information}, \penalty0 (14):\penalty0 208--209, 214, 2010.

\bibitem[He et~al.(2014)He, Li, Luo, et~al.]{he2014}
Qing He, Ning Li, Wenjuan Luo, et~al.
\newblock A review of machine learning algorithms under big data.
\newblock \emph{Pattern Recognition and Artificial Intelligence}, \penalty0 (4):\penalty0 327--336, 2014.

\bibitem[Liu et~al.(2017)Liu, He, Geng, et~al.]{liu2017}
Bin Liu, Jinrong He, Yaojun Geng, et~al.
\newblock Review of advances in basic systems for parallel machine learning algorithms.
\newblock \emph{Computer Engineering and Applications}, \penalty0 (11):\penalty0 31--38, 89, 2017.

\bibitem[Xiao(2017)]{xiao2017}
Hong Xiao.
\newblock Discussion on machine learning algorithms under big data.
\newblock \emph{Communication World}, \penalty0 (6):\penalty0 265--266, 2017.

\bibitem[Cao(2017)]{cao2017}
Xue Cao.
\newblock Machine learning: A catalyst for the ai revolution.
\newblock \emph{Electronics Technology and Software Engineering}, \penalty0 (13):\penalty0 255, 2017.

\bibitem[Lin and Chen(2017)]{lin2017}
Ling Lin and Fuji Chen.
\newblock Knowledge graph analysis of domestic network public opinion research based on citespace.
\newblock \emph{Information Science}, \penalty0 (2):\penalty0 119--125, 2017.

\bibitem[Wen and Wen(2016)]{wen2016}
Qianyu Wen and Tingxiao Wen.
\newblock Knowledge graph analysis of hospital management research based on citespace and ucinet.
\newblock \emph{Journal of Medical Informatics}, \penalty0 (8):\penalty0 70--75, 2016.

\bibitem[Chen(2016)]{chen2016}
Chen Chen.
\newblock \emph{Research on Analog Circuit Fault Diagnosis Based on Support Vector Machines}.
\newblock PhD thesis, Bohai University, Jinzhou, China, 2016.

\bibitem[Zhang et~al.(2017)Zhang, Guo, and Wang]{zhang2017}
Yi~Zhang, Quan Guo, and Jianyong Wang.
\newblock Neural network methods for big data analysis.
\newblock \emph{Engineering Science and Technology}, \penalty0 (1):\penalty0 9--18, 2017.

\bibitem[Jiang et~al.(2016)Jiang, Ding, Hou, et~al.]{jiang2016}
Junfeng Jiang, Xiangqian Ding, Ruichun Hou, et~al.
\newblock Visual analysis of big data research based on citespace iii.
\newblock \emph{Computer and Digital Engineering}, \penalty0 (2):\penalty0 291--295, 299, 2016.

\bibitem[Li et~al.(2014)Li, Zhang, and Xia]{li2014}
Deren Li, Liangpei Zhang, and Guisong Xia.
\newblock Automatic analysis and data mining of remote sensing big data.
\newblock \emph{Acta Geodaetica et Cartographica Sinica}, \penalty0 (12):\penalty0 1211--1216, 2014.

\end{thebibliography}

\end{document}